\documentclass{article}
\usepackage{spconf,amsmath,graphicx}
\usepackage{bm}
\usepackage{amssymb}
\usepackage[ruled,norelsize]{algorithm2e}
\usepackage{tabularx}

\makeatletter
\newcommand{\removelatexerror}{\let\@latex@error\@gobble}
\makeatother

\title{TENSOR-BASED NONLINEAR CLASSIFIER FOR HIGH-ORDER DATA ANALYSIS}
\name{K. Makantasis$^1$, A. Doulamis$^2$, N. Doulamis$^2$, A. Nikitakis$^3$, and A. Voulodimos$^2$$^{,4}$ \thanks{This paper is supported by the European Union Project TERPSICHORE funded under grant agreement 691218.}}
\address{$^1$ KIOS Research and Innovation Center of Excellence, University of Cyprus, Nicosia, Cyprus \\
$^2$ National Technical University of Athens, Athens, Greece \\
$^3$ Althexis Solutions Ltd, Nicosia, Cyprus \\
$^4$ Department of Informatics, Technological Educational Institute of Athens, Athens, Greece}
\begin{document}
%

\maketitle
\begin{abstract}
In this paper we propose a tensor-based nonlinear model for high-order data classification. The advantages of the proposed scheme are that (i) it significantly reduces the number of weight parameters, and hence of required training samples, and (ii) it retains the spatial structure of the input samples. The proposed model, called \textit{Rank}-1 FNN, is based on a modification of a feedforward neural network (FNN), such that its weights satisfy the {\it rank}-1 canonical decomposition. We also introduce a new learning algorithm to train the model, and we evaluate the \textit{Rank}-1 FNN on third-order hyperspectral data. Experimental results and comparisons indicate that the proposed model outperforms state of the art classification methods, including deep learning based ones, especially in cases with small numbers of available training samples.  
\end{abstract}
%

\section{Introduction}
\label{sec:intro}

Recent advances in sensing technologies have stimulated the development and deployment of sensors that can generate large amounts of high-order data. Interdependencies between information from different data modalities can improve the performance of data classification techniques \cite{zhou2016linked}. However, exploitation of high-order data raises new research challenges mainly due to the high dimensionality of the acquired information and, depending on the application at hand, the limited number of labeled examples \cite{camps2005kernel}.  

Tensor subspace learning methods, such as HOSVD, Tucker decomposition and CANDECOMP \cite{kolda2009tensor}, MPCA \cite{lu2008mpca} and probabilistic decompositions \cite{rai2014scalable,chu2009probabilistic,xu2015bayesian} have been proposed to tackle the dimensionality problem. These methods project the raw data to a lower dimensional space, in which the projected data can be considered as highly descriptive features of the raw information. The key problem in applying such methods in classifying high-order data, is that they do not take into consideration the data labels; 
therefore, the resulting features may not be sufficiently discriminative with respect to the classification task.
Tensor-based classifiers capable of mapping high-order data to desired outputs have also been proposed \cite{tan2012logistic,zhou2013tensor,li2014multilinear,tao2007general,hoff2015multilinear}. However, these methods are restricted to producing linear decision boundaries in feature space, and are therefore unable to cope with complex problems, where nonlinear decision boundaries are necessary to obtain classification results of high accuracy.
In order to better disentangle the input-output statistical relationships, deep learning approaches \cite{lecun1998gradient,bengio2007greedy} have been investigated for high-order data classification \cite{chen2014deep,Makantasis-etal:15,vakalopoulou2015building, makantasis2015deep}. Nevertheless, a typical deep learning architecture contains a huge number of tunable parameters, implying that a large number of labeled samples is also needed for accurate training. 


The present work draws its inspiration from  \cite{zhou2013tensor}, which proposes a linear tensor regression model for binary classification. In contrast to \cite{zhou2013tensor}, the paper at hand investigates a multi-class classification problem using a nonlinear tensor-based classifier. The proposed classifier is able to (i) handle raw high-order data without vectorizing them, and (ii) produce nonlinear decision boundaries, thus capturing complex statistical relationships between the data. 
The proposed scheme, henceforth called {\it Rank}-1 FNN, is based on a modification of a feedforward neural network (FNN), such that its weights satisfy the {\it rank}-1 canonical decomposition property, i.e., the weights are decomposed as a linear combination of a minimal number of possibly non-orthogonal {\it rank}-1 terms \cite{de2004computation}. Thence, the number of model parameters, and thus of training samples required, can be significantly reduced. We also introduce a new learning algorithm to train the network without violating the canonical decomposition property.

\section{Problem Formulation and Tensor algebra Notation}

\subsection{Problem Formulation}
\label{sec:problem_formulation}
Let us denote as $\bm X_i \in \mathbb R^{p_1 \times \cdots \times p_D}$ the $i$-th $D$-order tensor example that we aim at classifying into one of $C$ 
classes. Let us also denote as $p^k_w(\bm X_i)$ the probability of 
$\bm X_i$ belonging to the $k$-th class. 
Aggregating the values $p^k_w(\cdot)$ over all classes, we form a classification vector, $\bm y_i$, the elements of which $y_{i,k} \equiv p^k_w(\cdot)$. Then, the maximum $p^k_w(\cdot)$ value over all classes indicates the class to which the $\bm X_i$ 
belongs. 
The values of $y_{i,k}$ are estimated by minimizing a loss function over a dataset $\mathcal S = \{(\bm X_i, \bm t_i)\}_{i=1}^N$ during the training phase of a machine learning model. Vector $\bm t_i \in \{0,1\}^C$ and its elements $t_{i,j}$ are all zero except for one which equals unity indicating the class to which  $\bm X_i$ belongs. In the following, we omit subscript $i$ for simplicity purposes if we refer to an input sample.

\subsection{Tensor Algebra Notations and Definitions}
\label{sec:notation}
In this paper, tensors, vectors and scalars are denoted in bold uppercase, bold lowercase and lowercase letters, respectively. We hereby present some definitions that will be used through out this work.

\vspace{0.05in}
\noindent \textbf{Tensor vectorization}. The $vec(\bm B)$ operator stacks the entries of a $D$-order tensor $\bm B \in \mathbb R^{p_1 \times \cdots \times p_D}$ into a column vector. 

\vspace{0.05in}
\noindent \textbf{Tensor matricization}.The mode-\textit{d} matricization, $\bm B_{(d)}$, maps a tensor $\bm B$ into a $p_d \times \prod_{d' \neq d}p_{d'}$ matrix by arranging the mode-\textit{d} fibers to be the columns of the resulting matrix. 

\vspace{0.05in}
\noindent \textbf{\textit{Rank}-R decomposition}. A tensor $\bm B \in \mathbb R^{p_1 \times \cdots \times p_D}$ admits a \textit{rank}-R decomposition if  $\bm B = \sum_{r=1}^R \bm b_1^{(r)} \circ \cdots \circ \bm b_D^{(r)}$, where $\bm b_d^{(r)}\in \mathbb R^{p_d}$. The decomposition can be represented by $\bm B = [\![ \bm B_1, ... ,\bm B_D ]\!]$, where $\bm B_d = [\bm b_d^{(1)}, ... ,\bm b_d^{(R)}] \in \mathbb R^{p_d \times R}$. When a tensor $\bm B$ admits a \textit{rank}-R decomposition, it holds that:
\begin{equation}
\label{eq:3}
	\bm B_{(d)} = \bm B_d(\bm B_D \odot \cdots \odot \bm B_{d+1} \odot \bm B_{d-1} \odot \cdots \odot \bm B_1)^T
\end{equation}
where $\odot$ stands for the Khatri-Rao product. 
For more information on tensor algebra see \cite{kolda2009tensor}.

\section{High-order nonlinear modeling}
\label{sec:nonlinear}
The proposed \textit{Rank}-1 FNN is based on the concepts of \cite{zhou2013tensor}; however, in our case, the probability $p^k_w(\cdot)$ of an input example $\bm X$ belonging to the $k$-th class is nonlinearly interwoven with respect to the input tensor data and the weight parameters 
through a function $f_w(\cdot)$, i.e., $p^k_w(\bm X)=f_w(\bm X)$.
The main difficulty in implementing $p^k_w(\bm X)$ is that $f_w(\cdot)$ is actually unknown. One way to parameterize $f_w(\cdot)$ is to exploit the principles of the universal approximation theorem, stating that a function 
can be approximated by a FNN with a finite number of neurons within any degree of accuracy.

However, applying a FNN for high-order data classification involves two drawbacks. First, a large number of weights has to be learned; $Q\prod_{l=1}^D p_l + QC$, where $Q$ refers to the number of hidden neurons. 
This, in the sequel, implies that a large number of labeled samples are needed to successfully train the network. Second, the weights of the network are not directly related to the physical properties of the information belonging to different modes of the data, 
since the inputs are vectorized and thus they do not preserve their structure.

To overcome these problems, we propose a modification of FNN so that network weights from the input to the hidden layer satisfy the {\it rank}-1 canonical decomposition. 
Before presenting the {\it Rank}-1 FNN, we briefly describe how $p^k_w(\cdot)$ is modeled through a FNN.

\subsection{FNN Modeling}
\label{sec:FNN}
A FNN, with $Q$ hidden neurons, nonlinearly approximates the probability $p^k_w(\cdot)$ by associating a nonlinear activation function $g(\cdot)$ with each one of its hidden neurons. In this paper, the sigmoid function $g(x)=1/(1+\exp(-ax))$ is selected. The activation function of the $i$-th neuron receives as input the inner product of $vec(\bm X)$ and a weight vector $\bm w^{(i)}$ and produces as output a scalar $u_i$ given by
\begin{equation}
\label{eq:basic_model_in}
	u_i = g(\bm w^{(i)T} vec(X))\equiv g(\langle 	\bm w^{(i)}, \bm vec(X) \rangle).
\end{equation}
Gathering the responses of all hidden neurons in one vector $\bm u=[u_1, u_2,\cdots,u_Q]^T$, we have that
\begin{equation}
\label{eq:basic_model_in}
	\bm u = g(\langle \bm W, \bm X \rangle),
\end{equation}
where $\bm W=[\bm w^{(1)},\cdots, \bm w^{(Q)}]^T$ is a matrix containing the weights $\bm w^{(i)}$. 
Thus, the output of the network is given as  
\begin{equation}
\label{eq:basic_model}
	\bm p^{k}_w =\sigma ( \langle \bm v^{(k)},\bm u \rangle)\equiv \sigma (\bm v^{(k)T} \bm u), 
\end{equation}
where $\sigma(\cdot)$ stands for the softmax function, $\bm v^{(k)} $ the weights between the hidden and the output layer and the superscript for the $k$-th class.

\subsection{{\it Rank}-1 FNN Modeling}
To reduce the number of parameters of the network and to relate the classification results to the information belonging to different modes of the input data, we {\it rank}-1 canonically decompose the weight parameters $\bm w^{(i)}$ as:  
\begin{equation}
\label{eq:canonical2}
	\bm w^{(k)}=\bm w_D^{(k)} \otimes \cdots \otimes \bm w_1^{(k)}=\bm w_D^{(k)} \odot \cdots \odot \bm w_1^{(k)}.
\end{equation}
Eq. (\ref{eq:canonical2}) can be seen as an expression of the Khatri-Rao product, which is the column-wise Kronecker product, denoted as $\otimes$, of the {\it rank}-1 canonical decomposition weight parameters $\bm w_l^{(k)}$. Thus, $\bm w_l^{(k)} \in \mathbb R^{p_l}$ and the total number of \textit{Rank}-1 FNN is $Q\sum_{l=1}^D p_l + QC$. Based on the statements of Section \ref{sec:notation}, it holds that
\begin{equation}
\label{eq:observation1}
\begin{split}
	\langle \bm w_D^{(k)} \odot \cdots \odot \bm w_1^{(k)},& \bm X \rangle = 
	 \langle \bm w_l^{(k)}, \bm X_{(l)}(\bm w_D^{(k)}\odot \cdots \\ &\odot \bm w_{l+1}^{(k)}\odot \bm w_{l-1}^{(k)}\odot \cdots \odot \bm w_1^{(k)}) \rangle.
\end{split}
\end{equation}
In Eq. (\ref{eq:observation1}), $\bm X_{(l)}$ denotes the mode-$l$ matricization of tensor $\bm X$.
Then, taking into account the properties of Eq. (\ref{eq:observation1}), the output of the $i$-th hidden neuron $u_i$ can be written as 
\begin{equation}
\label{eq:basic_model_in_2}
\begin{split}
	u_i & = g(\langle \bm w^{(i)}, \bm X \rangle) 
	=g(\langle \bm w_D^{(i)} \otimes \cdots \otimes \bm w_1^{(i)}, \bm X \rangle) \\
	&= g(\langle \bm w_D^{(i)} \odot \cdots \odot \bm w_1^{(i)}, \bm X \rangle) 
	= g(\langle \bm w_l^{(i)}, \bm \tau_{\neq l}^{(i)} \rangle).
\end{split}
\end{equation} 
Vector $\bm \tau_{\neq l}^{(i)}$ is a transformed version of input $\bm X$, that is,
\begin{equation}
	\bm \tau_{\neq l}^{(i)} = \bm X_{(l)}(\bm w_D^{(k)}\odot \cdots \odot \bm w_{l+1}^{(k)}\odot \bm w_{l-1}^{(k)}\odot \cdots \odot \bm w_1^{(k)})
\end{equation}
and is independent from $\bm w_l^{(i)}$. 
Eq. (\ref{eq:basic_model_in_2}) actually resembles the operation of a single perceptron having as inputs the weights $\bm w_l^{(i)}$ and the transformed version $\tau_{\neq l}$ of the input data. In other words, if the {\it rank}-1 canonically decomposed weights $\bm w_r^{(i)}$ with $r\neq l $ are known, then $\tau_{\neq l}^{(i)}$ will be also known. 
The main modification of this structure compared to a typical FNN lies in the hidden layer, where the weights of a hidden neuron are first decomposed into $D$ canonical factors.

\subsection{The Learning Algorithm}
\label{sec:learning_nonlinear}
Let us aggregate the total \textit{Rank}-1 FNN weight parameters as 
\begin{equation}
\label{eq:aggregate_weights}
\bm W_l = [\bm w_l^{(1)} \bm w_l^{(2)} \cdots \bm w_l^{(Q)}], \bm V = [\bm v^{(1)} \bm v^{(2)} \cdots \bm v^{(C)}]
\end{equation} 
with $l=1,2,\cdots,D$. In order to train the proposed model a set $\mathcal S = \{ (\bm X_i, \bm t_i) \}_{i=1}^N$ is used. The learning algorithm minimizes the negative log-likelihood 
\begin{equation}
\label{eq:neg_log_likelihood}
    L(\bm W_1,...,\bm W_D, \bm V ;\mathcal S) = -\sum_{i=1}^N \sum_{k=1}^{C} t_{i,k}\log p^k_w(\bm X_i),
\end{equation}
with respect to network responses $\bm y_i=[\cdots y_{i,k}\cdots]^T$, with $y_{i,k}\equiv p_w^k(\bm X_i)$, and targets $\bm t_i$ over all training samples.


The weights of the \textit{Rank}-1 FNN must satisfy the {\it rank}-1 canonical decomposition expressed by Eq. (\ref{eq:canonical2}). Assuming that all weights $\bm V$ and $\bm w_r^{(i)}$ with $r\neq l$ are fixed, vector $\bm \tau_{\neq l}^{(i)}$ can be estimated; therefore, vector $\bm w_l^{(i)}$ is the only unknown parameter of the network. This vector can be derived through a gradient based optimization algorithm, assuming that the derivative $\partial L / \partial \bm w_l^{(i)}$ is known. This derivative can be computed using the backpropagation algorithm. Therefore, an estimation of the parameters of the {\it Rank}-1 FNN is obtained by iteratively solving with respect to one of the $D$ canonical decomposed weight vectors, assuming the remaining fixed. Algorithm \ref{alg:2} presents the steps of the proposed algorithm.

\begin{figure}[t]
{\begingroup
\removelatexerror
\begin{algorithm}[H]
\caption{Estimation of of the {\it Rank}-1 FNN Weights}
\label{alg:2}
\SetAlgoLined
 \textbf{Initialization:}\\
 1. Set Iteration Index $n\rightarrow 0$\\
 2. Randomize all the weight $\bm w_l^{(i)}(n)$ and $\bm v^{(k)}(n)$ \\
 for $l=1,...,D$, $i=1,2,\cdots,Q$, $k=1,...,C$  \\
 3. \Repeat{termination criteria are met}
                 {
  \For{$l=1,...,D$} {
            \For {$i=1,...Q$}
              {
                 3.1 Estimate the transformed input vector 
                 $\tau_{\neq l}^{(i)}= \bm X_{(l)}(\bm w_D^{(i)}(n)\odot \cdots \odot \bm w_{l+1}^{(i)}(n)\odot \bm w_{l-1}^{(i)}(n+1)\odot \cdots \odot \bm w_1^{(i)}(n+1)),$ \\
                 3.2 Update the weights $\bm w_l^{(i)}(n)$ towards the negative direction of $\partial L / \partial \bm w_l^{(i)}$ 
              }
   
  	}
  	\For{$k=1,...,C$} {
    3.3 Update the weights $\bm v^{(k)}(n)$ towards the negative direction of $\partial E / \partial \bm v^{(k)}$  
  	}
  	Set $n \rightarrow n+1$
  }
\end{algorithm}
\endgroup}
\end{figure}

\begin{figure}[t]
  \begin{minipage}{0.9\linewidth}
    \centering
    \centerline{\includegraphics[height=5.25cm]{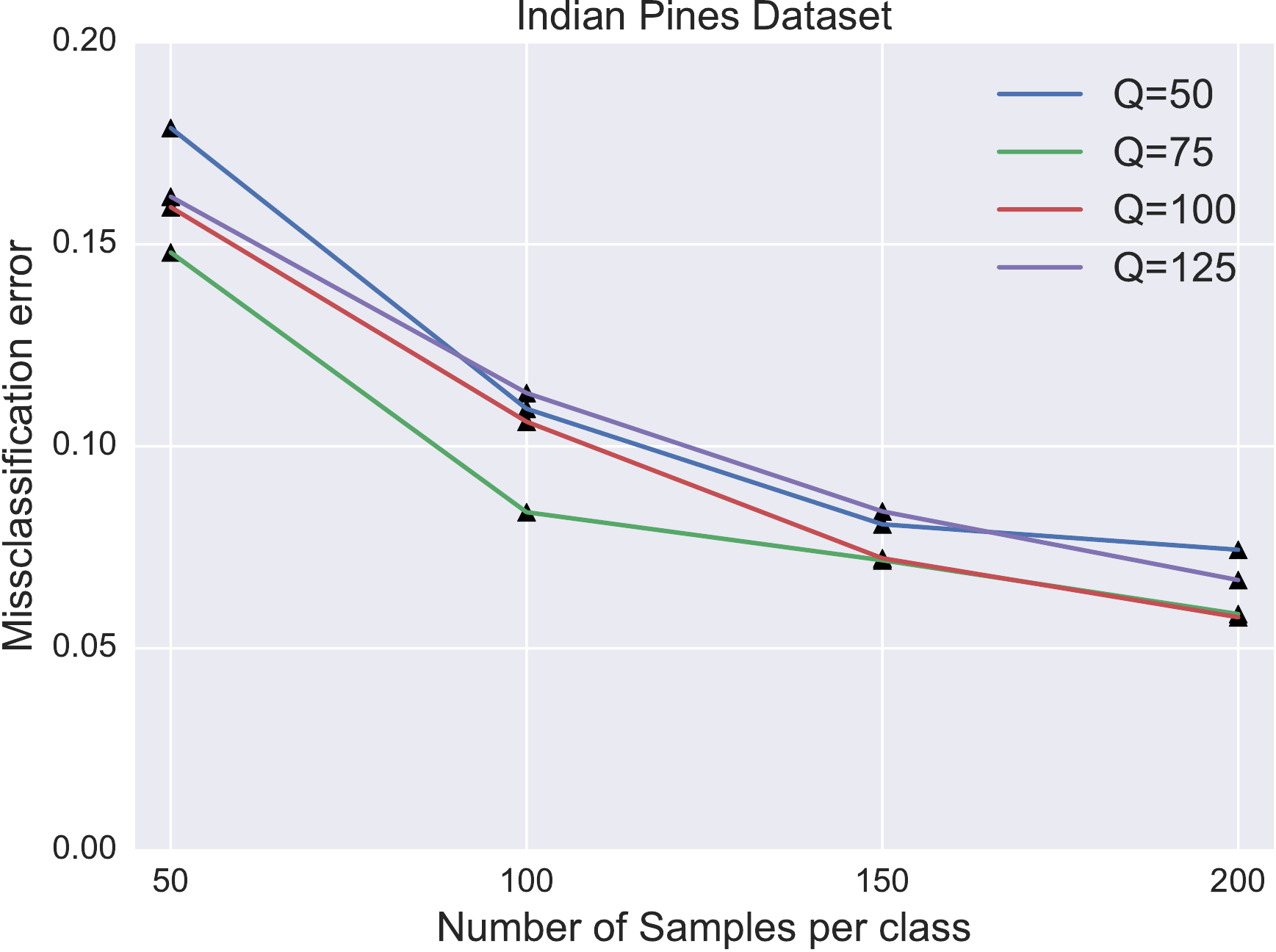}}
  \end{minipage} 
  \begin{minipage}{0.9\linewidth}
    \centering
    \centerline{\includegraphics[height=5.25cm]{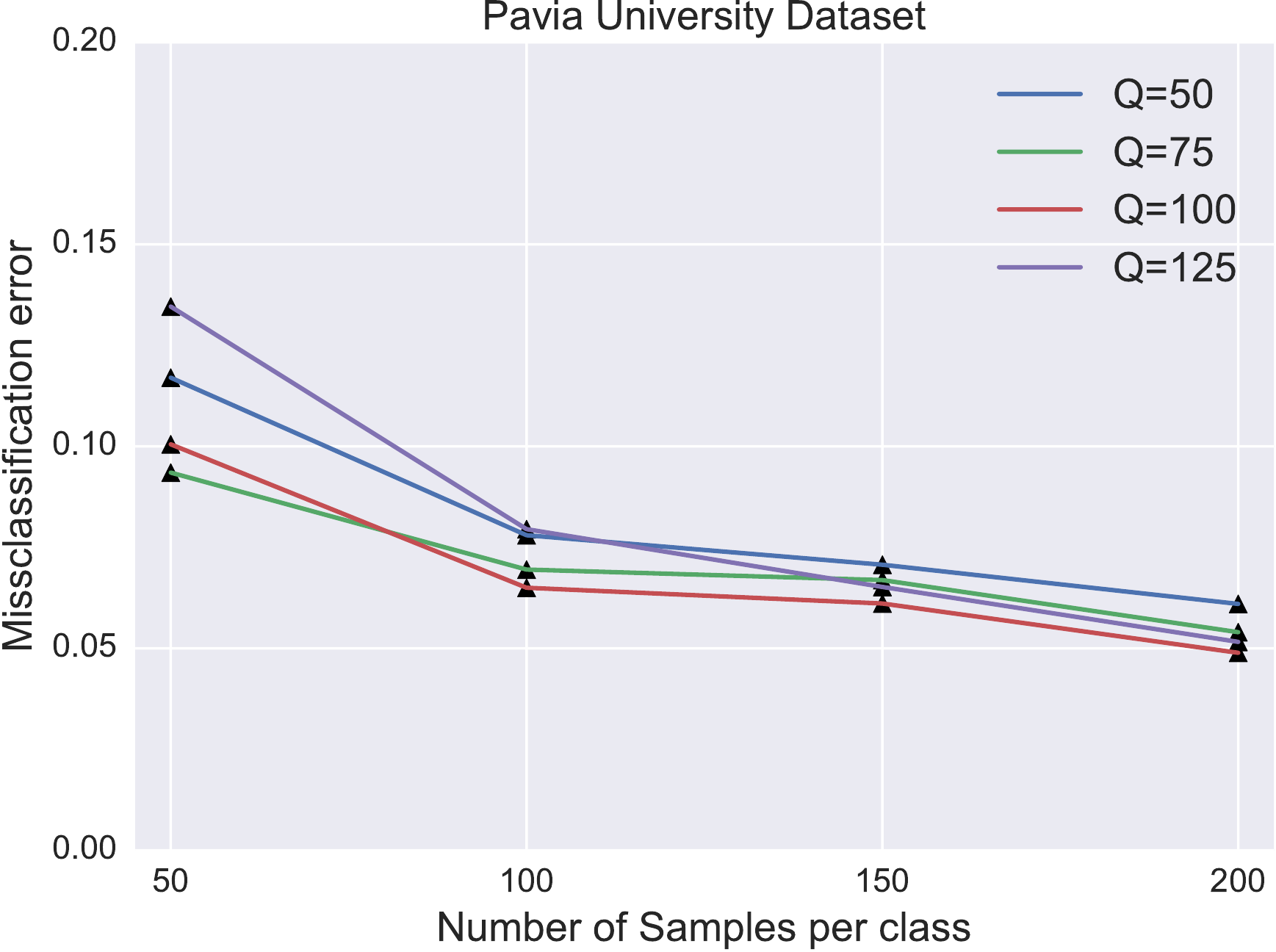}}
  \end{minipage}
  \caption{Misclassification error on test set versus the complexity, determined by $Q$, of the \textit{Rank}-1 FNN.}
  \label{fig:2}
\end{figure}

\section{EVALUATION ON HYPERSPECTRAL DATA}
\label{sec:experiments}
To investigate whether the reduced number of parameters would limit the descriptive power of the \textit{Rank}-1 FNN, we conduct experiments and present quantitative results regarding its classification accuracy on 3-order hyperspectral data. 
In our study, we used (i) the Indian Pines dataset \cite{baumgardner2015220}, which consists of 224 spectral bands and $10,086$ labeled pixels and (ii) the Pavia University dataset \cite{plaza2009recent}, consisting of 103 spectral bands and $42,776$ labeled pixels. 

A hyperspectral image is represented as a 3-order tensor of dimensions $p_1 \times p_1 \times p_3$, where $p_1$ and $p_2$ correspond to the height and width of the image and $p_3$ to the spectral bands. In order to classify a pixel $I_{x,y}$ at location $(x,y)$ on image plane and fuse spectral and spatial information, we use a square patch of size $s \times s$ centered at $(x,y)$. Let us denote as ${\bm t}_{x,y}$ the label of $I_{x,y}$ and as ${\bm X}_{x,y}$ the tensor patch centered at $(x,y)$. Then, we can form a dataset $S=\{({\bm X}_{x,y}, {\bm t}_{x,y}) \}$, which is used to train the classifier. 

To evaluate the performance of the {\it Rank}-1 FNN, we conducted different experiments using training datasets of 50, 100, 150 and 200 samples from each class respectively. Initially, we evaluate the performance of the \textit{Rank}-1 FNN with respect to its complexity, i.e., the value of $Q$, indicating the number of hidden neurons. Particularly, we set $Q$ to be equal to 50, 75, 100 and 125. Greater values of $Q$ imply a more complex model. The results of this evaluation are presented in Fig.\ref{fig:2}.
Regarding the Indian Pines dataset, we observe that the model with $Q=75$ outperforms all other models. When the training set size is very small, i.e. 50 samples per class, the model with $Q=50$ underfits the data. On the other hand, the models with $Q=100$ and $Q=125$  slightly overfit the data due to their high complexity.
As far as the Pavia University dataset is concerned, we observe that the model with $Q=100$ outperforms all other models, when the dataset size is larger than 50 samples per class. When the training dataset size is 50 samples per class the model with $Q=75$ outperforms all other models. The model with $Q=125$ overfits the data, while the model with $Q=50$ underfits them. In both datasets, as training set increases, the misclassification error decreases. 

In the following, we compare the performance of \textit{Rank}-1 FNN against FNN, RBF-SVM, and two deep learning approaches that have been proposed for classifying hyperspectral data; the first one is based on Stacked-Autoencoders (SAE) \cite{chen2014deep}, while the second on the exploitation of Convolutional Neural Networks (CNN) \cite{Makantasis-etal:15}. The FNN consists of one hidden layer with 75 hidden neurons when trained on Indian Pines dataset and 100 hidden neurons for the Pavia University dataset (as derived from Fig.\ref{fig:2}). The architecture of the network based on SAE consists of three hidden layers, while each hidden layer contains 10\% less neurons than its input. The number of hidden neurons from one hidden layer to the next is gradually reduced, so as not to allow the network to learn the identity function during pre-training. Regarding CNN, we utilize exactly the same architecture as the one presented in \cite{Makantasis-etal:15}. The performance of all these models is evaluated on varying size training sets.

\begin{table}[t]
\centering
\caption{Classification accuracy results (\%) of \textit{Rank}-1 FNN}
\newcolumntype{L}[1]{>{\hsize=#1\hsize\raggedright\arraybackslash}X}%
\newcolumntype{C}[1]{>{\hsize=#1\hsize\centering\arraybackslash}X}%
\label{table:2}

\begin{tabularx}{0.98\linewidth}{L{8.8}C{2.4}C{2.4}C{2.4}C{2.4}}
\hline \hline 
\multicolumn{5}{c}{{\bf Pavia University}} \\ \hline 
Samples per class & 50 & 100 & 150 & 200 \\ \hline

{\it Rank}-1 FNN (Q=100)  & \textbf{89.95} & \textbf{93.50} & 93.89 & 95.11   \\ \hline
FCFFNN        		  & 67.79 & 76.53 & 78.48 & 82.59   \\ \hline
RBF-SVM       		  & 86.98 & 88.99 & 89.86 & 91.82   \\ \hline 
SAE           		  & 86.54 & 91.90 & 92.38 & 93.29   \\ \hline 
CNN           		  & 88.89 & 92.74 & \textbf{94.68} & \textbf{95.89}   \\ \hline \hline

\multicolumn{5}{c}{{\bf Indian Pines}} \\ \hline 
Samples per class & 50 & 100 & 150 & 200 \\ \hline

{\it Rank}-1 FNN (Q=75)  & \textbf{85.20} & \textbf{91.63} & \textbf{92.82} & 94.15   \\ \hline
FCFFNN        		 & 73.88 & 81.10 & 84.14 & 85.86   \\ \hline
RBF-SVM       		 & 73.18 & 77.86 & 82.11 & 84.99   \\ \hline 
SAE           		 & 65.51 & 70.66 & 74.03 & 76.49   \\ \hline 
CNN           		 & 82.43 & 85.48 & 92.28 & \textbf{94.81}   \\ \hline \hline

\end{tabularx}
\end{table}

Table \ref{table:2} presents the outcome of this comparison. When the training set size is small, our approach outperforms all other models. This stems from the fact that the proposed \textit{Rank}-1 FNN exploits tensor algebra operations to reduce the number of coefficients that need to be estimated during training, while at the same time it is able to retain the spatial structure of the input. Although the FNN utilizes the same number of hidden neurons as our proposed model, it seems to overfit training sets when a small size dataset is used, due to the fact that it employs a larger number of coefficients. RBF-SVM performs better than the FNN on the Pavia University dataset, but slightly worse on the Indian Pines dataset. 
The full connectivity property of SAE implies very high complexity, which is responsible for its poor performance, due to overfitting in the Indian Pines dataset. Finally, the CNN-based approach performs better than FNN, RBF-SVM and SAE mainly because of its sparse connectivity (low complexity) and the fact that it can exploit the spatial information of the input. When the training set consists of 150 and 200 samples per class, for the Pavia University dataset, and 200 samples for the Indian Pines dataset, the CNN-based approach seems to even outperform the \textit{Rank}-1 FNN. This happens because the CNN-based model has higher capacity than the proposed model, which implies that it is capable of better capturing the statistical relationships between the input and the output, when the training set contains sufficient information. However, when the size of the training set is small, which is often the case, \textit{Rank}-1 FNN, due to its lower complexity, consistently outperforms the CNN-based model.

\section{Conclusions}
\label{sec:conclusions}
In this work, we present a nonlinear tensor-based scheme for high-order data classification. The proposed model is characterized by (i) the small number of weight parameters and (ii) its ability to retain the spatial structure of the high-order input samples. 
We have evaluated the performance of the model on 3-order hyperspectral data in terms of classification accuracy by comparing it against other nonlinear classifiers, including state-of-the-art  deep  learning  models.  The  results  indicate that  in cases where  the  size  of  the  training  set  is  small,  the proposed \textit{Rank}-1 FNN  presents  superior  performance  against the  compared methods, including deep learning based ones.

\bibliographystyle{IEEEbib}
\bibliography{refs}

\end{document}